%% file: main.tex
\documentclass[10pt,twocolumn,letterpaper]{article}

\usepackage{cvpr}              

\input{preamble}

\definecolor{cvprblue}{rgb}{0.21,0.49,0.74}
\usepackage[pagebackref,breaklinks,colorlinks,citecolor=cvprblue]{hyperref}
\usepackage{multirow}
\usepackage{colortbl}
\usepackage{pifont}
\usepackage{marvosym}

\title{Attribute-Aware Representation Rectification for Generalized Zero-Shot Learning}
\author{
\textbf{Zhijie Rao\textsuperscript{\rm 1}}, 
\textbf{Jingcai Guo\textsuperscript{\rm 1,\rm 2,\Letter}\thanks{Jingcai Guo is the corresponding author.}},
\textbf{Xiaocheng Lu\textsuperscript{\rm 4}}, 
\textbf{Qihua Zhou\textsuperscript{\rm 1}}, \\
\textbf{Jie Zhang\textsuperscript{\rm 1}}, 
\textbf{Kang Wei\textsuperscript{\rm 1}}, 
\textbf{Chenxin Li\textsuperscript{\rm 3}}, 
~and \textbf{Song Guo\textsuperscript{\rm 4}}\\
\textsuperscript{\rm 1}Department of Computing, The Hong Kong Polytechnic University, Hong Kong SAR\\
\textsuperscript{\rm 2}Hong Kong Polytechnic University Shenzhen Research Institute, China\\
\textsuperscript{\rm 3}Department of Electronic Engineering, The Chinese University of Hong Kong, Hong Kong SAR\\
\textsuperscript{\rm 4}Department of CSE, Hong Kong University of Science and Technology, Hong Kong SAR\\
{\tt\small\{zhijie.rao,jc-jingcai.guo,qi-hua.zhou,jie-comp.zhang,adam-kang.wei\}@polyu.edu.hk}\\
{\tt\small\{xiaocheng.lu,songguo\}@cse.ust.hk, chenxinli@link.cuhk.edu.hk}
}

\begin{document}
\maketitle
\input{sec/0_abstract}
\input{sec/1_intro}
\input{sec/2_related}
\input{sec/3_method}
\input{sec/4_experiment}
\input{sec/5_conclusion}

{
    \small
    \bibliographystyle{ieeenat_fullname}
    \bibliography{main}
}


\end{document}

%% file: preamble.tex
\usepackage[dvipsnames]{xcolor}


%% file: sec/0_abstract.tex
\begin{abstract}
Generalized Zero-shot Learning (GZSL) has yielded remarkable performance by designing a series of unbiased visual-semantics mappings, wherein, the precision relies heavily on the completeness of extracted visual features from both seen and unseen classes. 
However, as a common practice in GZSL, the pre-trained feature extractor may easily exhibit difficulty in capturing domain-specific traits of the downstream tasks/datasets to provide fine-grained discriminative features, i.e., domain bias, which hinders the overall recognition performance, especially for unseen classes. 
Recent studies partially address this issue by fine-tuning feature extractors, while may inevitably incur catastrophic forgetting and overfitting issues. 
In this paper, we propose a simple yet effective \textbf{\underline{A}}ttribute-\textbf{\underline{A}}ware \textbf{\underline{R}}epresentation \textbf{\underline{R}}ectification framework for GZSL, dubbed 
$\mathbf{(AR)^{2}}$, 
to adaptively rectify the feature extractor to learn novel features while keeping original valuable features. 
Specifically, our method consists of two key components, i.e., Unseen-Aware Distillation (UAD) and Attribute-Guided Learning (AGL). 
During training, 
UAD exploits the prior knowledge of attribute texts that are shared by both seen/unseen classes with attention mechanisms to detect and maintain unseen class-sensitive visual features in a targeted manner, 
and meanwhile, 
AGL aims to steer the model to focus on valuable features and suppress them to fit noisy elements in the seen classes by attribute-guided representation learning. 
Extensive experiments on various benchmark datasets demonstrate the effectiveness of our method~\footnote{The code is available at: \url{github.com/zjrao/AARR}}.
\end{abstract}

%% file: sec/1_intro.tex
\section{Introduction}
\label{sec:intro}

\begin{figure}[htbp]
    \centering
    \begin{minipage}{\linewidth}
    \includegraphics[width=0.99\textwidth]{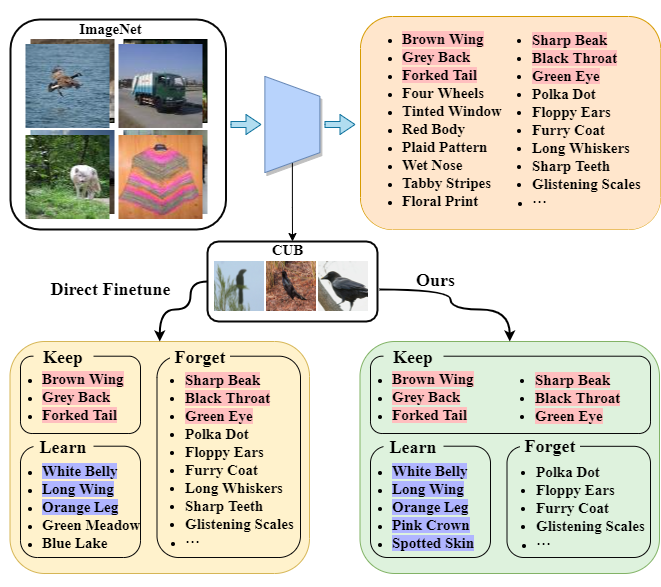}
    \end{minipage}
    \caption{\textit{Pink} indicates features that are beneficial to the downstream data. \textit{Purple} indicates newly learned valuable features. Pre-trained models lack the ability to extract fine-grained information of specific datasets. Direct fine-tuning leads to forgetfulness and overfitting, whereas our approach aims to keep and learn valuable information (better viewed in color).}
    \label{fig:intro}
\end{figure}

The success of deep neural networks in the visual area is well documented, and one of the reasons is the support of a huge amount of training data. 
Nevertheless, it is an intractable condition to fulfill in the real-world scenario, e.g., there are very few or even no samples of rare species. To solve this problem, Zero-Shot Learning (ZSL) has gained increasing attention in recent years with its dedication to bridging the gap between language and vision, which takes inspiration from the logical reasoning ability of human beings. By virtue of the shared visual information and the semantic descriptions of unseen classes, ZSL endows the model with out-of-distribution recognition capability. Another more challenging task is Generalized Zero-Shot Learning (GZSL), which requires simultaneous recognition of both seen and unseen categories \cite{lampert2009learning, chao2016empirical}.

Mainstream studies strive to search and locate local visual attributes so as to construct one-to-one visual-semantic mappings \cite{xie2019attentive, xie2020region, chen2022transzero, xu2022vgse, guo2023graph, schonfeld2019generalized, chen2022msdn}, or implicitly learn the class-wise visual-semantic relation to simulate unseen distributions in a generative manner \cite{xian2018feature, verma2018generalized, felix2018multi, li2019leveraging, yu2020episode, keshari2020generalized, vyas2020leveraging, xie2022leveraging}. Despite the promising results achieved by these approaches, the visual-semantic matchiness is significantly restricted by the incompleteness of the visual features. Such deficiency stems from the domain bias between the pre-trained dataset, e.g., ImageNet \cite{deng2009imagenet}, and the downstream tasks/datasets, which simply states that the extracted features are insufficient to provide fine-grained knowledge to build complete and reliable visual-semantic pairs \cite{chen2021free}.

To address such issues, a straightforward solution is to fine-tune them on downstream tasks, which may inevitably introduce catastrophic forgetting and overfitting problems (Fig. \ref{fig:intro}) \cite{hendrycks2019using,li2019rethinking}. 
Concretely, deep neural networks usually excel at seeking shortcuts from observed data \cite{geirhos2020shortcut}, favoring partial features that benefit the seen classes during the fine-tuning process, while conversely, features that are critical for the unseen classes may be filtered. 
Meanwhile, newly learned features may contain various noise, e.g., backgrounds, irrelevant attributes, etc., rendering the model to create pseudo-visual-semantic associations. 
Worse still, identical issues also exist in the realm of continual learning \cite{de2021continual} and spawn substantial mature schemes. However, they are not applicable to ZSL/GZSL, which has no further training or adjustment phase for novel categories. 

Grounded on the above discussions, we lock our research interests in the following two aspects: 1) \textit{How to keep the valuable knowledge in the raw features to ensure the unseen distributional generalizability of the model}, and 2) \textit{How to guide the learning process of new knowledge to reduce the interference of noisy factors}. 
The crux of these issues is to determine which features are worth being kept and learned. In this paper, we argue that attributes, i.e., textual descriptions or semantic embeddings, are the only shared supervised signals to distinguish seen and unseen classes. Thus, the kept and learned features need to be directed by them.

Hereby, we present the \textbf{\underline{A}}ttribute-\textbf{\underline{A}}ware \textbf{\underline{R}}epresentation \textbf{\underline{R}}ectification framework for GZSL, i.e., $\mathbf{(AR)^{2}}$, to adaptively rectify the feature extractor to learn novel features while keeping original valuable features. 
Concretely, our $\mathbf{(AR)^{2}}$ consists of two key components including Unseen-Aware Distillation (UAD) and Attribute-Guided Learning (AGL). The objective of UAD is to identify those in the raw features that are beneficial to both seen and unseen classes. 
By virtue of the attention mechanism of the class activation map \cite{selvaraju2017grad} and the attribute-region classifier \cite{huynh2020fine, chen2022msdn}, specific features are identified and localized. 
Specifically, on the one hand, we use the pre-trained network as the teacher model and attribute labels of similar classes as supervisory information to obtain class activation maps of unseen classes by gradient propagation. Meanwhile, the score of the attribute-region classifier is utilized to restrict the scope of attention maps, from which valuable features are filtered out. In parallel, the student model is prompted to retain this fraction of features by means of feature distillation. 
On the other hand, AGL aims to encourage the model to refine features that are most relevant to attributes and reduce noisy interference. We first leverage features extracted by the teacher model to initialize visual prototypes of each sub-attribute, which form an attribute pool. 
Class prototypes can then be obtained by selecting and assembling various sub-attributes from the pool. The attribute pool is updated by each batch of data during the training period, with increasing the semantic distance between prototypes of each category as the learning goal. In this way, the model is implicitly motivated to learn features that are associated with attributes and are discriminative. Finally, the teacher model is updated by the exponential moving average method with the student model.

Our contributions are summarized as follows:

\begin{itemize}
    \item To alleviate the issue of mismatch between the pre-trained model and downstream data/tasks, we present a novel method named $(AR)^{2}$ to adaptively rectify the feature representations. $(AR)^{2}$ steers the learning process with the supervision of attributes, continuously keeping and learning the most valuable knowledge.

    \item $(AR)^{2}$ consists of two main components, wherein, UAD assists the model in reviewing old knowledge to prevent catastrophic forgetting and AGL guides the model in refining features to avoid overfitting on noisy information.

    \item We conduct extensive experiments and analysis on three benchmark datasets, and the results show that the proposed method is effective to improve the performance of the model.
    
\end{itemize}

%% file: sec/2_related.tex
\section{Related Work}
\label{sec:related}

\subsection{Bridge Vision and Attribute}

ZSL/GZSL aims to learn shared attributes (semantics) from accessible training data, thereby obtaining the ability to infer on the unknown domain \cite{lampert2009learning, chao2016empirical}. Attribute descriptions, i.e., text embeddings, category prototypes, etc., are the only prior information to access unseen categories. Therefore, how to connect the visual features of the seen classes with the attributes is a central issue in ZSL/GZSL. Numerous studies choose the most direct way, i.e., mapping visual features to the attribute space \cite{romera2015embarrassingly}. In this direction, the accuracy of the mapping is the main challenge to attack. Existing approaches include generating hallucinatory categories \cite{changpinyo2016synthesized}, reconstructing visual features \cite{kodirov2017semantic}, and modeling region-attribute relationships \cite{xie2019attentive, liu2020attribute, huynh2020fine, xie2020region, xu2020attribute, wang2021dual, chen2022transzero, xu2022vgse, guo2023graph, Li2023boosting}, to name a few. In contrast, some studies opt to map attributes to the visual space and adjust the mapping function by maintaining semantic relations among categories \cite{zhang2017learning}. In addition to this, some studies combine the characteristics of the above two methods by mapping visual features and attributes to a common space \cite{liu2018generalized, jiang2019transferable, schonfeld2019generalized, chen2022msdn}, thus mitigating the discrepancies between different modal data. Compared to modeling the visual-attribute relationship explicitly, generative approaches \cite{xian2018feature, verma2018generalized, felix2018multi, li2019leveraging, yu2020episode, keshari2020generalized, vyas2020leveraging, xie2022leveraging} provide an alternative perspective, i.e., learning the relationship implicitly by means of the distributional alignment capability of GANs or VAEs. Although these approaches achieve promising results, they are limited by the domain bias problem between the pre-trained model and the downstream dataset, i.e., the pre-trained model struggles to capture the fine-grained features of the specific dataset, which is detrimental to the establishment of accurate visual-attribute associations.

\subsection{Domain Bias}

A common practice in ZSL/GZSL is to utilize the ImageNet~ \cite{deng2009imagenet} pre-trained model to extract features and then develop links between visual features and attributes. However, inherent domain differences exist between datasets, i.e., domain bias. For instance, ImageNet \cite{deng2009imagenet} lacks fine-grained annotations for birds, which are needed to discriminate samples in the CUB dataset \cite{wah2011caltech}. Xian \textit{et al.} \cite{xian2019f} achieve performance improvement by fine-tuning the feature extractor, demonstrating the existence of domain bias. However, they do not conduct further research to mitigate the forgetting and overfitting problems \cite{hendrycks2019using,li2019rethinking}.

\subsection{Representation Rectification}

Domain bias leads to features extracted by the pre-trained model on the downstream dataset being incomplete, i.e., lacking fine-grained, targeted information. To this end, some studies attempt to rectify the extracted features (raw features). Li \textit{et al.} \cite{li2021generalized} and Chen \textit{et al.} \cite{chen2021semantics} argue that the raw features contain both class-relevant and class-irrelevant parts, which are then stripped away by means of disentanglement. Chen \textit{et al.} \cite{chen2021free} strive to refine the raw features in order to reduce the redundancy of the information and to enhance the discriminability of the features. Han \textit{et al.} \cite{han2021contrastive}, on the other hand, utilize contrastive learning to bring the same-class representations closer and push the dissimilarity of representations farther away. Kong \textit{et al.} \cite{kong2022compactness} then resort to enhancing the intra-class compactness. Although their methods mitigate the domain bias to some extent, they are unable to learn new knowledge on downstream datasets.

%% file: sec/3_method.tex
\section{Method}
\label{sec:method}

\begin{figure*}[htbp]
    \centering
    \begin{minipage}{\linewidth}
    \includegraphics[width=0.99\textwidth]{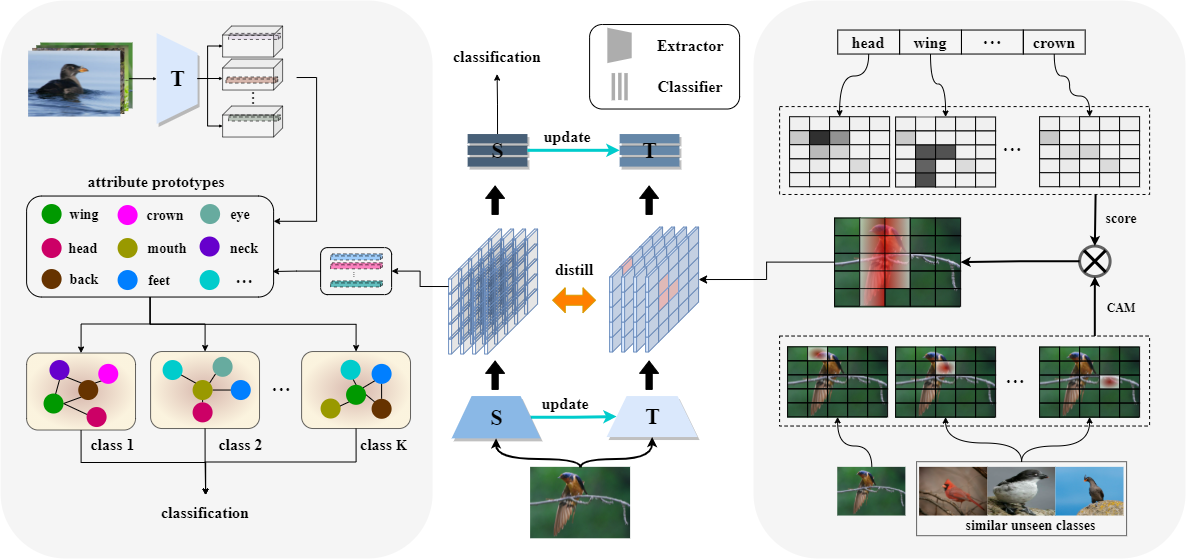}
    \end{minipage}
    \caption{The overall framework of our method. `S' denotes the student model and `T' denotes the teacher model. UAD identifies and localizes the features extracted by the teacher model to filter the valuable parts. AGL utilizes the features extracted by the student model to update the attribute pool to facilitate the associations between the learned features and attributes (better viewed in color).}
    \label{fig:method}
\end{figure*}

\noindent\textbf{Preliminary.} In the ZSL/GZSL tasks, generally, the training data consists of complete samples of seen classes and attribute vectors of unseen classes. Suppose $D^s=\{x^s, y^s, a^s\}$ denotes the seen class sample set, where $x^s, y^s$ and $a^s$ denote the images, labels and attributes, respectively. Let $D^u=\{a^u\}$ denote attributes of unseen classes, and $a=a^s\cup a^u$ is the whole attribute set. Meanwhile, we also use the semantic embeddings of each sub-attribute learned by GloVe, which are indicated by $v=\{v_1, v_2, ..., v_n\}$, where $n$ is the number of sub-attributes. Let $K$ denote the number of categories, including seen and unseen categories. Our approach consists of a teacher model $\mathcal{T}=\{\mathcal{E}^t, \mathcal{W}^t\}$ and a student model $\mathcal{S}=\{\mathcal{E}^s, \mathcal{W}^s\}$ that have the same network structure, where $\mathcal{W}^t=[\mathcal{W}_1^t, \mathcal{W}_2^t]$ and $\mathcal{W}^s=[\mathcal{W}_1^s, \mathcal{W}_2^s]$. Each model includes a feature extractor $\mathcal{E}$ and a classifier $\mathcal{W}=[\mathcal{W}_1, \mathcal{W}_2]$. Let $w_1, w_2$ denote the learnable parameters of $\mathcal{W}_1, \mathcal{W}_2$, respectively. Then $w_1^t, w_2^t$ stand for the teacher model's parameters, and $w_1^s, w_2^s$ represent the student model's. AGL module has a classifier $\mathcal{W}^p$, where $w^p$ denotes the parameters.

\noindent\textbf{Overview.} Our approach is depicted in Fig. \ref{fig:method}. The teacher model is responsible for retaining historical features to prevent the student model from forgetting old knowledge. UAD employs a dual filtering mechanism, i.e., class activation map and attribute scoring, to identify and locate the valuable parts of the features extracted by the teacher model. The output of UAD is a weight map, which measures the value of each region. After that, the teacher's knowledge is passed on to the students via distillation. AGL utilizes the original and learnable features to maintain a pool of attributes that contain prototypes of each sub-attribute. Different class prototypes are available by selecting and assembling various sub-attributes. By maximizing the distance between the prototypes of each class, AGL implicitly facilitates the correlation between learned features and attributes.

\subsection{Attribute-Region Classifier}

Attributes are the only shared prior information between seen and unseen classes, which are crucial for identifying unknown classes. However, attribute descriptions are typically coarse-grained class-wise annotations, hence many studies resort to learning fine-grained attribute-region mapping relations. Recent works \cite{huynh2020fine, chen2022msdn} have achieved promising results by incorporating attention mechanisms into classifiers. Since our approach requires localizing attribute-related features (regions), we employ the same classifier.

Assume $f=\mathcal{E}(x), f\in\mathbb{R}^{CHW}$ denotes the feature of an input image extracted by the feature extractor $\mathcal{E}$, where $C, H, W$ denote the channel, height, and width. Then $f$ is divided into $r=HW$ regions, and each region is represented by a $C$-dimensional vector. The degree of association between each attribute and each region can be scored, which is formulated as: 
\begin{equation}
    p(i, j | w_1) = \frac{\exp{(v_iw_1f_j)}}{\sum_{q=1}^r\exp{(v_iw_1f_q)}},\label{eq:p(i,j)}
\end{equation}
where $p(i, j)$ denotes the association score of attribute $i$ and region $j$. Then $p$ is used to weight the final output and optimize it with cross-entropy loss. The loss function is defined as:
\begin{equation}
    \mathcal{L}_{CE}(a, f, p | w_2)=-\log\frac{\exp{(a(vw_2f)p)}}{\sum^K_{k=1}\exp{(a_k(vw_2f)p)}},
\end{equation}

\subsection{Unseen-Aware Distillation}

Models gradually forget some important features as they learn downstream data. In order to help the model memorize these features, we design the UAD module to review the old valuable knowledge. A key issue is how to recognize which features are worth being retained. We are inspired by the class activation map \cite{selvaraju2017grad}, which indicates the correlation between regional features and classes via gradient responses. So we use the class activation maps created by the teacher model to represent the importance of each region. Let $f_t=\mathcal{E}^t(x)$ denote the feature extracted by the teacher extractor, and $o_t=\mathcal{W}^t(f_t)$ denotes the output of the teacher classifier. Then, if we want to know which regional features are strongly correlated with category $c$, we just need to activate the gradient for the corresponding category and the gradient map is represented as:
\begin{equation}
    g(k)=\tau(\frac{\partial \mathcal{L}_{CE}(a_c, f_t, p_t | w_2^t)}{\partial f_t}),
\end{equation}
where $a_c$ is the attribute of class $c$ and $p_t(i, j)=p(i, j | w_1^t)$ and $\tau(\cdot)$ is Min-Max Normalization.

\noindent\textbf{Unseen-Aware Attention.} A simple idea is to activate the category corresponding to the training sample to get the activation map. However, such an approach would be inclined to focus on attributes that favor the seen classes, leading to overfitting, which is not conducive to generalization to the unseen classes. Therefore, we need to know which attributes the unseen classes are interested in. In the end, we activate similar unseen classes at the same time. Specifically, we first compute the similarity between the attributes of the unseen and seen classes, which is measured by Euclidean distance. Then we select the $m$ most similar seen classes for each unseen class, corresponding to the fact that there will exist an unseen class similarity set $U_k$ for each seen class $k$. After that, we get the new activation map:
\begin{equation}
    g=\tau(g(k)+\frac{1}{d}\sum_{k_u\in U_k}g(k_u)),
\end{equation}
where $d$ denotes the size of set $U_k$.

\noindent\textbf{Attribute-Aware Attention.} Despite the fact that the class activation map implies connections between regions and classes, those regions may be the ones that contain noise, such as backgrounds. For example, if the unseen class possesses the attribute \textit{crown} while the corresponding seen class does not, then the activated region is inaccurate. To suppress the effect of this part of the regions, we reweight the activation map with the score of the attribute-region classifier. Assuming $\hat{p}$ denotes the score map of the training sample computed by Eq. \ref{eq:p(i,j)}, the final attention map is defined as:
\begin{equation}
    g = g\cdot \hat{p},
\end{equation}
where $(\cdot)$ denotes the dot product.

\noindent\textbf{Knowledge Distillation.} We use feature distillation to transfer knowledge to the student model. Suppose $f_s=\mathcal{E}^s(x)$ denotes the feature extracted by the student extractor. The distillation loss is:
\begin{equation}
    \mathcal{L}_{UAD}=||f_s-f_t||^2\cdot g,
\end{equation}
where $||\cdot||$ means Mean-Square Error loss function.

\subsection{Attribute-Guided Learning}

Another problem with the application of the model to downstream tasks is overfitting. Due to the characteristic of neural networks that are skilled at finding shortcuts \cite{geirhos2020shortcut}, some noisy features, irrelevant attributes, etc. in the seen classes receive more attention, which prevents the model from generalizing to the unseen classes. The goal of AGL is to guide the model to learn features that are relevant to attributes. Our motivation is to reorganize learned features into class prototypes guided by attributes, and then implicitly enhance the connection between learned features and attributes by increasing the distinguishability of class prototypes.

\noindent\textbf{Initialize Attribute Pool.} We firstly create an attribute pool $h=\{h_1, h_2, ..., h_n\}, h_i\in \mathbb{R}^C$, where $n$ is the number of sub-attribute and $h_i$ denotes the prototypical feature of the $i$-th sub-attribute. The attribute pool is initialized by the features extracted by the teacher model. Specifically, we compute the prototypes using the region in each sample that has the highest correlation with the attribute. According to Eq. \ref{eq:p(i,j)}, we can obtain the association score map of attributes and regions. Then the prototype of attribute $i$ is formulated as:
\begin{equation}
    h_i = \frac{\sum_{j=1}^{N}\Bar{p}_i^j\Bar{f}_i^j}{\sum_{j=1}^{N}\Bar{p}_i^j},
\end{equation}
where $\Bar{p}_i^j$ denotes the max score of attribute $i$ in sample $j$ and $\Bar{f}_i^j$ is the corresponding region feature of $\Bar{p}_i^j$. $N$ is the size of the whole training dataset. Note that the initialization is performed only once during the entire training process, and we set $h$ learnable.

\noindent\textbf{Update Attribute Pool.} During the training phase, we update the attribute pool with the features extracted from the student model. For a batch of features $f_s$ extracted by the student extractor, the prototype of attribute $i$ is:
\begin{equation}
    \Bar{h}_i=\frac{1}{B}\sum_{b=1}^B\sum_{j=1}^{r} \frac{p^b(i, j)f_j^b}{p^b(i, j)},
\end{equation}
where $\Bar{h}_i$ denotes the prototype of attribute $i$ computed by current batch and $B$ denotes the batch size. Here $p(i,j)=p(i,j|w_1^s)$. Then the attribute pool is updated by:
\begin{equation}
    h = h\times\lambda + \Bar{h}\times(1.0-\lambda),\label{eq:lambda}
\end{equation}
where $\lambda$ is a balanced parameter and we set it learnable.

\noindent\textbf{Optimization Objective.} We hope that the updated attribute pool can facilitate the recognition of both seen and unseen classes. Specifically, with the help of attribute vectors, class prototypes are obtained by adaptively selecting and assembling sub-attributes. Then we increase the semantic distance between the class prototypes to enhance the correlation between the learned features and attributes. The loss function is:
\begin{equation}
    \mathcal{L}_{AGL}=-\log\frac{\exp{(ahw_p)}}{\sum^K_{k=1}\exp{(a_khw_p)}}.
\end{equation}

\subsection{Overall Objective}

In the pre-training stage, only $\mathcal{L}_{CE}$ is used for training because the performance of the attribute-region classifier is too weak to localize valuable features. When the model is stabilized, all loss functions are used to train together. The optimization objective is:
\begin{equation}
    \mathcal{L}_{AR} = \mathcal{L}_{CE} + \beta\mathcal{L}_{UAD} + \gamma\mathcal{L}_{AGL},
\end{equation}
where $\beta$ and $\gamma$ are hyper-parameters. The teacher model does not participate in training. At the end of each epoch, an update is performed by the exponential moving average method. Let $\Theta^t$ and $\Theta^s$ denote the parameters of the teacher model and the student model. The teacher model is updated by:
\begin{equation}
    \Theta^t = \Theta^t\times\delta+\Theta^s\times(1.0-\delta),
\end{equation}
where $\delta$ is a constant and is set to $0.9995$.

%% file: sec/4_experiment.tex
\section{Experiments}
\label{sec:experiments}

\begin{table*}[]
    \centering
    \caption{The experimental results(\%) of CUB, SUN, and AWA2 on \textbf{ZSL} and \textbf{GZSL} settings. Method types are listed in the \emph{BRANCH} column. GEN: Generative Method; OWM: One-Way Mapping; CS: Common Space; DA: Data Augmentation; and RR: Representation Rectification. The best, second-best, and third-best results are highlighted in \textcolor{red}{\textbf{red}}, \textcolor{blue}{\textbf{blue}}, and \underline{underlined}, respectively.}
    \centering
    \resizebox{\textwidth}{!}{
    \begin{tabular}{l c c c c c c c c c c c c c}
        \toprule
         & & \multicolumn{4}{c}{\textbf{CUB}} & \multicolumn{4}{c}{\textbf{SUN}} & \multicolumn{4}{c}{\textbf{AWA2}} \\
        \cmidrule(r){3-6} \cmidrule(r){7-10} \cmidrule(r){11-14}
        METHOD & BRANCH & T & U & S & H & T & U & S & H & T & U & S & H\\
        \hline
        f-CLSWGAN (CVPR $'18$) \cite{xian2018feature} & GEN & 57.3 & 43.7 & 57.7 & 49.7 & 60.8 & 42.6 & 36.6 & 39.4 & 68.2 & 57.9 & 61.4 & 59.6\\
        f-VAEGAN-D2 (CVPR $'19$) \cite{xian2019f} & GEN & 61.0 & 48.4 & 60.1 & 53.6 & 64.7 & 45.1 & 38.0 & 41.3 & 71.1 & 57.6 & 70.6 & 63.5\\
        TF-VAEGAN (ECCV $'20$) \cite{narayan2020latent} & GEN & 64.9 & 52.8 & 64.7 & 58.1 & \textcolor{blue}{\textbf{66.0}} & 45.6 & \textcolor{red}{\textbf{40.7}} & 43.0 & \textcolor{blue}{\textbf{72.2}} & 59.8 & 75.1 & 66.6\\
        E-PGN (CVPR $'20$) \cite{yu2020episode} & GEN & 72.4 & 52.0 & 61.1 & 56.2 & - & - & - & - & \textcolor{red}{\textbf{73.4}} & 52.6 & 83.5 & 64.6\\
        SGMA (NeurIPS $'19$) \cite{zhu2019semantic} & OWM & 71.0 & 36.7 & 71.3 & 48.5 & - & - & - & - & 68.8 & 37.6 & 87.1 & 52.5\\
        AREN (CVPR $'19$) \cite{xie2019attentive} & OWM & 71.8 & 38.9 & \textcolor{blue}{\textbf{78.7}} & 52.1 & 60.6 & 19.0 & 38.8 & 25.5 & 67.9 & 15.6 & \textcolor{blue}{\textbf{92.9}} & 26.7\\
        LFGAA (ICCV $'19$) \cite{liu2019attribute} & OWM & 67.6 & 36.2 & \textcolor{red}{\textbf{80.9}} & 50.0 & 61.5 & 18.5 & \textcolor{blue}{\textbf{40.0}} & 25.3 & 68.1 & 27.0 & \textcolor{red}{\textbf{93.4}} & 41.9\\
        DAZLE (CVPR $'20$) \cite{huynh2020fine} & OWM & 66.0 & 56.7 & 59.6 & 58.1 & 59.4 & \textcolor{red}{\textbf{52.3}} & 24.3 & 33.2 & 67.9 & 60.3 & 75.7 & 67.1\\
        APN (NeurIPS $'20$) \cite{xu2020attribute} & OWM & 72.0 & 65.3 & 69.3 & 67.2 & 61.6 & 41.9 & 34.0 & 37.6 & 68.4 & 57.1 & 72.4 & 63.9\\
        VS-Boost (IJCAI $'23$) \cite{Li2023boosting} & OWM & \textcolor{blue}{\textbf{79.8}} & 68.0 & 68.7 & \underline{68.4} & 62.4 & 49.2 & 37.4 & 42.5 & - & - & - & -\\
        DCN (NeurIPS $'18$) \cite{liu2018generalized} & CS & 56.2 & 28.4 & 60.7 & 38.7 & 61.8 & 25.5 & 37.0 & 30.2 & 65.2 & 25.5 & 84.2 & 39.1\\
        CADA-VAE (CVPR $'19$) \cite{schonfeld2019generalized} & CS & 59.8 & 51.6 & 53.5 & 52.4 & 61.7 & 47.2 & 35.7 & 40.6 & 63.0 & 55.8 & 75.0 & 63.9\\
        HSVA (NeurIPS $'21$) \cite{chen2021hsva} & CS & 62.8 & 52.7 & 58.3 & 55.3 & 63.8 & 48.6 & \underline{39.0} & \textcolor{blue}{\textbf{43.3}} & - & 59.3 & 76.6 & 66.8\\
        MSDN (CVPR $'22$) \cite{chen2022msdn} & CS & 76.1 & \underline{68.7} & 67.5 & 68.1 & \underline{65.8} & \textcolor{blue}{\textbf{52.2}} & 34.2 & 41.3 & 70.1 & \underline{62.0} & 74.5 & 67.7\\
        HAS (MM $'23$) \cite{chen2023zero} & DA & 76.5 & \textcolor{blue}{\textbf{69.6}} & \underline{74.1} & \textcolor{blue}{\textbf{71.8}} & 63.2 & 42.8 & 38.9 & 40.8 & 71.4 & \textcolor{blue}{\textbf{63.1}} & \underline{87.3} & \textcolor{red}{\textbf{73.3}}\\
        FREE (ICCV $'21$) \cite{chen2021free} & RR & - & 55.7 & 59.9 & 57.7 & - & 47.4 & 37.2 & 41.7 & - & 60.4 & 75.4 & 67.1\\
        SDGZSL (ICCV $'21$) \cite{chen2021semantics} & RR & 75.5 & 59.9 & 66.4 & 63.0 & 62.4 & 48.2 & 36.1 & 41.3 & \underline{72.1} & \textcolor{red}{\textbf{64.6}} & 73.6 & \underline{68.8}\\
        CE-GZSL (CVPR $'21$) \cite{han2021contrastive} & RR & \underline{77.5} & 63.9 & 66.8 & 65.3 & 63.3 & 48.8 & 38.6 & \underline{43.1} & 70.4 & \textcolor{blue}{\textbf{63.1}} & 78.6 & \textcolor{blue}{\textbf{70.0}}\\
        \rowcolor{yellow!17}
        $\mathbf{(AR)^{2}}$(\textbf{Ours}) & RR & \textcolor{red}{\textbf{80.2}} & \textcolor{red}{\textbf{74.1}} & 73.0 & \textcolor{red}{\textbf{73.5}} & \textcolor{red}{\textbf{66.2}} & \underline{51.9} & 37.8 & \textcolor{red}{\textbf{43.7}} & 70.9 & 61.5 & 81.2 & \textcolor{blue}{\textbf{70.0}}\\ 
        \bottomrule
    \end{tabular}}
    \label{tab:mainresults}
\end{table*}

\begin{table}[]
    \centering
    \caption{The results(\%) of ablation study. `*' denotes MSDN \cite{chen2022msdn} as the baseline. `\ding{51}' denotes adding the module. The best results are marked in \textbf{bold}.}
    \centering
    \resizebox{0.48\textwidth}{!}{
    \begin{tabular}{lcccccccc}
         \toprule
         & & & \multicolumn{2}{c}{\textbf{CUB}} & \multicolumn{2}{c}{\textbf{SUN}} & \multicolumn{2}{c}{\textbf{AWA2}}\\
         \cmidrule(r){4-5}\cmidrule(r){6-7}\cmidrule(r){8-9}
         METHOD & UAD & AGL & T & H & T & H & T & H\\
         \hline
         Baseline* & & & 76.1 & 68.1 & 65.8 & 41.3 & 70.1 & 67.7\\
         Finetune & & & 78.5 & 71.8 & 63.8 & 42.0 & 68.4 & 67.3 \\
         Ours-1 & \ding{51} & & 79.7 & 73.0 & 65.4 & 42.9 & 69.5 & 68.4\\
         Ours-2 &  & \ding{51} & 79.1 & 72.2 & 64.5 & 42.1 & 68.3 & 67.5\\
         \rowcolor{yellow!17} Ours & \ding{51} & \ding{51}& \textbf{80.2} & \textbf{73.5} & \textbf{66.2} & \textbf{43.7} & \textbf{70.9} & \textbf{70.0}\\
         \bottomrule
    \end{tabular}}
    \label{tab:ablation}
\end{table}

\noindent\textbf{Datasets.} We perform experiments on three benchmark datasets including CUB (Caltech UCSD Birds
200) \cite{wah2011caltech}, SUN (SUN Attribute) \cite{patterson2012sun}, and AWA2 (Animals
with Attributes 2) \cite{xian2017zero}. We split the seen and unseen classes according to the criteria described in \cite{xian2017zero}. CUB is a fine-grained bird dataset containing 11,788 images with 150 seen classes and 50 unseen classes, and the attribute dimension is 312. SUN is a scene dataset containing 14,340 images with 645 seen classes and 72 unseen classes, and the attribute dimension is 102. AWA2 is a coarse-grained animal dataset containing 37,322 images, including 40 seen classes and 10 unseen classes, and the attribute dimension is 85.

\noindent\textbf{Evaluation Protocols.} For the \textbf{ZSL setting}, we evaluate the top-1 accuracy on unseen classes and denote it as \textbf{T}. For the \textbf{GZSL setting}, we record the top-1 accuracies on seen and unseen classes and denote them as \textbf{S} and \textbf{U}, respectively. Meanwhile, we report their harmonic mean \textbf{H}, i.e., $\mathrm{H}=\frac{2\mathrm{S}\mathrm{U}}{\mathrm{S}+\mathrm{U}}$, to evaluate the performance of GZSL.

\noindent\textbf{Implementation Details.} We adopt the feature extractor of ResNet101 \cite{he2016deep} pre-trained on ImageNet \cite{deng2009imagenet} and the classifier of MSDN \cite{chen2022msdn} to form our network architecture. The batch size is set to 32 for CUB and 50 for SUN and AWA2. We set the learning rate to 5e-6 and employ the RMSProp optimizer with the momentum set as 0.9 and weight decay set as 1e-4. For hyperparameters, we set $m$ to 5 for AWA2 and 10 for CUB and SUN. For $\beta$ and $\gamma$, we set them to \{10, 0.1\} for CUB and AWA2 and \{15, 0.1\} for SUN.

\subsection{Comparision with State-of-the-Arts}

\begin{figure*}[htbp]
    \centering
    \begin{minipage}{\linewidth}
    \includegraphics[width=0.99\textwidth]{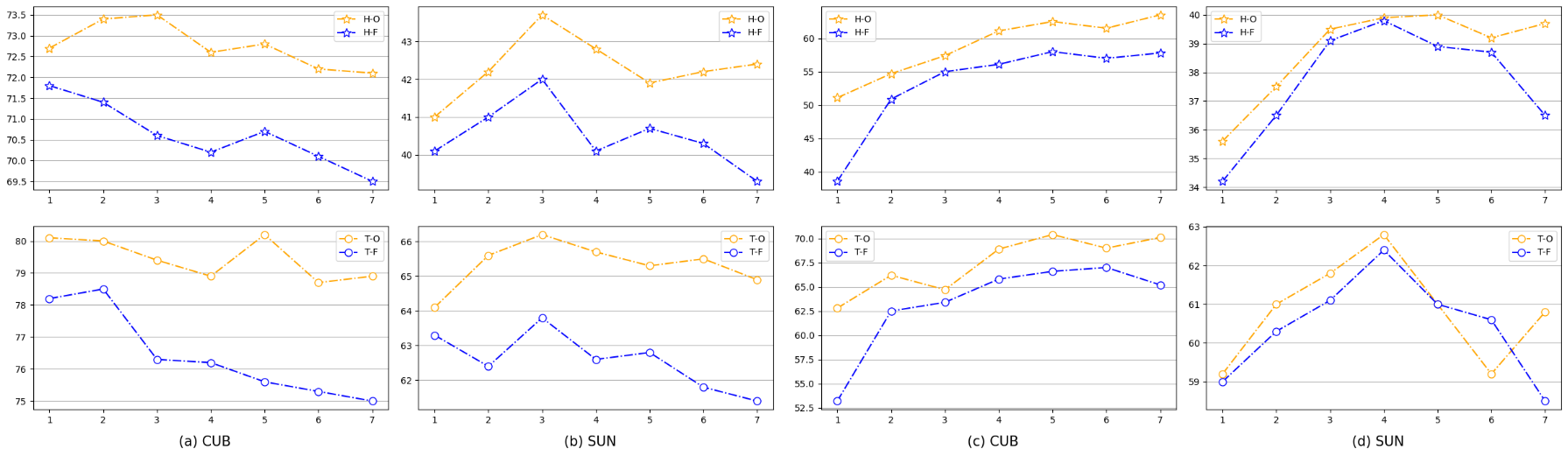}
    \end{minipage}
    \caption{Comparison of the stability of our method with finetune. \emph{H-O}: H score of our method. \emph{H-F}: H score of Finetune. \emph{T-O}: T score of our method. \emph{T-F}: T score of Finetune (better viewed in color).}
    \label{fig:exper_stability}
\end{figure*}

\begin{figure}[htbp]
    \centering
    \begin{minipage}{\linewidth}
    \includegraphics[width=0.99\textwidth]{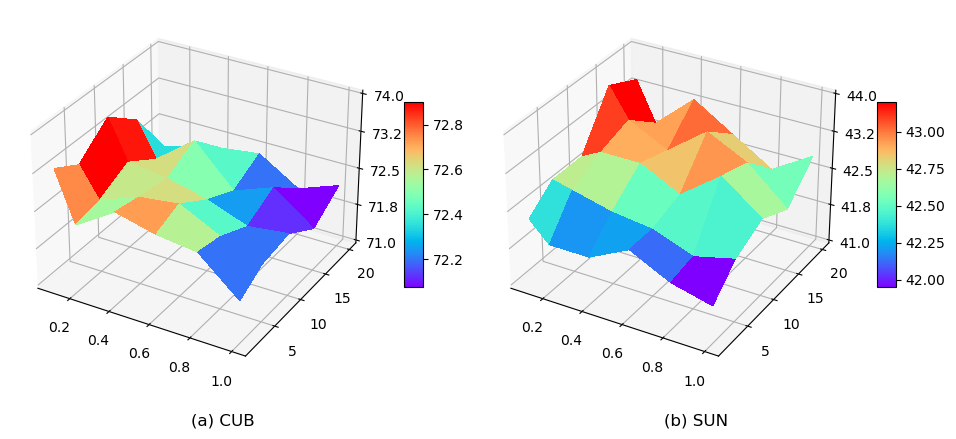}
    \end{minipage}
    \caption{The analysis of sensitivity to hyperparameters $\beta$ and $\gamma$ (better viewed in color).}
    \label{fig:exper_sense_of_hyper}
\end{figure}

We compare our $\mathbf{(AR)^{2}}$ with state-of-the-art methods of various types, including GEN (Generative Method), OWM (One-Way Mapping, i.e., visual feature maps to attribute space or attribute maps to feature space), CS (Common Space), RR (Representation Rectification). The experimental results are shown in Table \ref{tab:mainresults}. As can be seen from the table, our method yields competitive results, with the highest H-scores and ZSL accuracies on both the CUB and SUN datasets. Specifically, our H-score on CUB precedes the second place by 1.7\%, and the recognition rate of unseen classes (74.1\%) precedes the second place by 4.5\%. On the SUN dataset, we achieve an H-score of 43.7\% and a T-score of 66.2\% for the first and third places, respectively. Five of our metrics achieve the best scores, one second and one third. Notably, our scheme is also far ahead of its peers. The experimental results show that the features extracted from the pre-trained model have great room for improvement and that representation rectification is one of the effective schemes. Meanwhile, our solution significantly contributes to the recognition performance with dual constraints of keeping and learning valuable knowledge.

\subsection{Ablation Study}

We perform a series of ablation experiments to analyze the functionality of each component. We use MSDN as the baseline and compare our method with direct fine-tuning. The results of the experiments are shown in Table \ref{tab:ablation}, where it can be seen that fine-tuning brings some boost on the CUB and SUN datasets, but does not significantly promote the effect on AWA2. Our method, instead, achieves the best results on all three datasets, and each component plays a positive role. It demonstrates the soundness of the design of each component and the effectiveness of our method in mitigating the forgetting and overfitting problems.

\subsection{Analysis of Hyperparameters}

\noindent\textbf{Sensitivity of $\beta$ and $\gamma$.} We conduct a number of experiments to analyze the sensitivity of the hyperparameters $\beta$ and $\gamma$. The experimental results are shown in Fig. \ref{fig:exper_sense_of_hyper}. We set $\beta$ to [1, 5, 10, 15, 20] and $\gamma$ to [0.1, 0.2, 0.4, 0.6, 0.8, 1.0], respectively. Fig. \ref{fig:exper_sense_of_hyper} (a) and (b) show the performance plots of the H-score on CUB and SUN, respectively. As can be seen from the figure, changes in $\beta$ and $\gamma$ do not lead to large fluctuations in performance. And, we recommend setting $\beta$ to [10, 15] and $\gamma$ below $0.5$ for optimal performance.

\noindent\textbf{Effect of $m$ and $\lambda$.} We further analyze the impact of the parameter $m$, i.e., how many of the most similar seen classes are appropriate to choose among the UDA module. The experimental results are shown in Table \ref{tab:m_and_lamda}. We set $m$ to [1, 5, 10] for comparison, and the experimental results on CUB show that setting it to 10 is best, but setting it to 5 is better on AWA2. The reason is that CUB has a total of 200 classes with 150 seen classes, while AWA2 has a total of only 50 classes with 40 seen classes. Moreover, AWA2 is a coarse-grained dataset with less similarity between categories. In addition, we investigate the effect of the parameter $\lambda$ in Eq. \ref{eq:lambda}. The experimental results are also shown in Table \ref{tab:m_and_lamda}. We conduct experiments on the CUB and SUN datasets by fixing it to [0.9, 0.5]. The results show that the fixed values are not as effective as the learnable ones.

\begin{table}[htbp]
    \centering
    \caption{The effect of the number $m$ of similar seen classes and the parameter $\lambda$ in Eq. (\ref{eq:lambda}).}
    \centering
    \begin{tabular}{lcccc}
         \toprule
         & \multicolumn{2}{c}{\textbf{CUB}} & \multicolumn{2}{c}{\textbf{AWA2}}\\
         \cmidrule(r){2-3}\cmidrule(r){4-5}
         SETTING & T & H & T & H \\
         \hline
         $m=1$ & 79.6 & 72.1 & 69.6 & 69.0 \\
         $m=5$ & 79.5 & 73.0 & 70.9 & 70.0 \\
         $m=10$ & 80.2 & 73.5 & 69.1 & 68.7 \\
         \toprule
         & \multicolumn{2}{c}{\textbf{CUB}} & \multicolumn{2}{c}{\textbf{SUN}}\\
         \cmidrule(r){2-3}\cmidrule(r){4-5}
         SETTING & T & H & T & H \\
         \hline
         $\lambda=0.9$ & 78.4 & 72.5 & 64.9 & 43.0 \\
         $\lambda=0.5$ & 79.6 & 72.9 & 64.5 & 42.8 \\
         $\lambda=learnable$ & 80.2 & 73.5 & 66.2 & 43.7 \\
         \bottomrule
    \end{tabular}
    \label{tab:m_and_lamda}
\end{table}

\begin{figure*}[htbp]
    \centering
    \begin{minipage}{\linewidth}
    \includegraphics[width=0.99\textwidth]{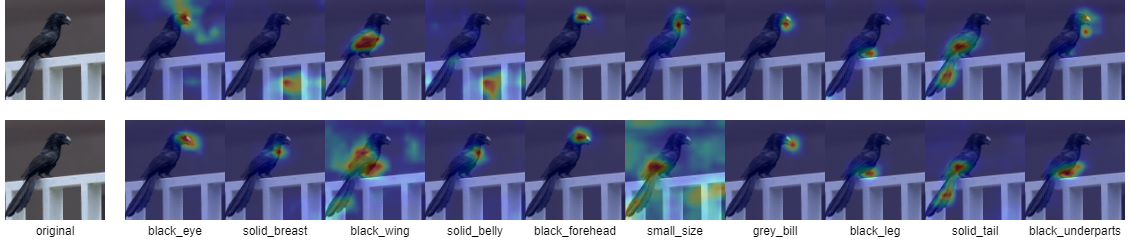}
    \end{minipage}
    \caption{Visualization of the attention heat map. The first row represents the heat map of MSDN and the second row denotes our method (better viewed in color).}
    \label{fig:exper_attri_attention}
\end{figure*}

\begin{figure*}[htbp]
    \centering
    \begin{minipage}{\linewidth}
    \includegraphics[width=0.99\textwidth]{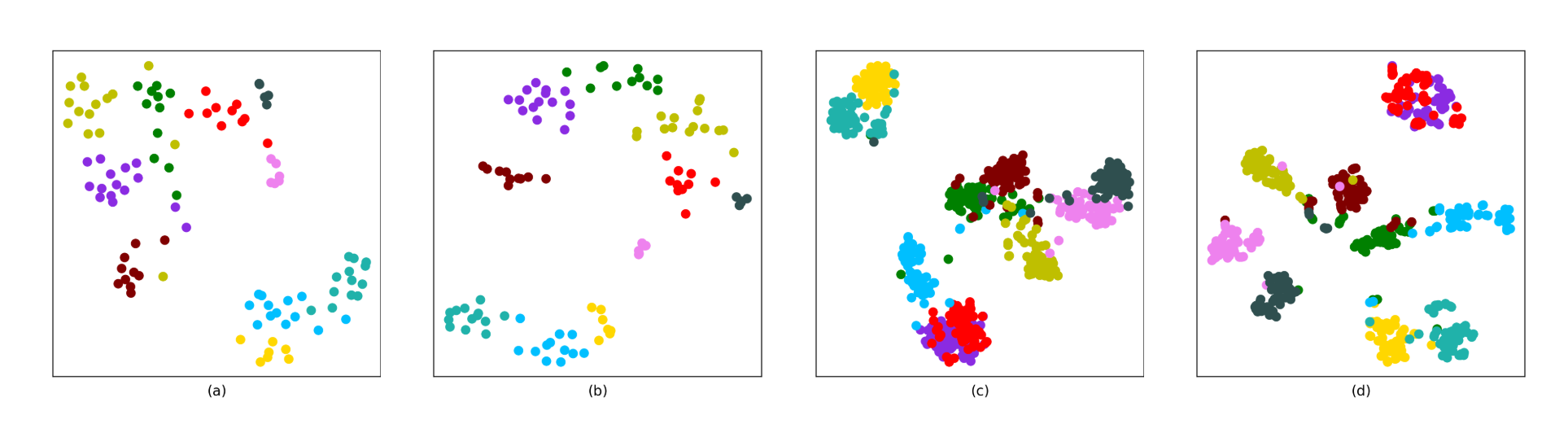}
    \end{minipage}
    \caption{The t-SNE visualization of our method and MSDN. (a-b): seen class features; (c-d): unseen class features. (a, c): MSDN, (b, d): Our method (better viewed in color).}
    \label{fig:tsne}
\end{figure*}

\subsection{Stability Analysis}

We conduct a plethora of experiments to analyze the stability of our method and compare it with fine-tuning. We study the effect of training time as well as learning rate on performance to analyze the contribution of our method in mitigating forgetting and overfitting. The experimental results are shown in Fig. \ref{fig:exper_stability}. In the figure, the horizontal axis represents the training time, i.e., the training epoch. Fig. \ref{fig:exper_stability} (a) and (b) represent the results with the learning rate as 5e-6, Fig. \ref{fig:exper_stability} (c) represents the results with the learning rate as 5e-5, and Fig. \ref{fig:exper_stability} (d) represents the results with the learning rate as 1e-5. From the figures, it can be seen that our method is relatively more stable, especially the results in Fig. \ref{fig:exper_stability} (a) and Fig. \ref{fig:exper_stability} (b), which show that our method can effectively suppress the forgetting problem of the model. In Fig. \ref{fig:exper_stability} (c) and Fig. \ref{fig:exper_stability} (d), there are fluctuations caused by the excessive learning rate, but our method can adjust quickly and obtain higher performance than fine-tuning. It indicates that our AGL module effectively captures the attribute-related features and boosts the performance.

\subsection{Visualization of Attribute-Region Attention}

In order to investigate the features that our scheme keeps and learns during the learning process, we perform a visualization analysis. As shown in Fig. \ref{fig:exper_attri_attention}, the first row represents the baseline method, and the second row is our method. It can be seen that MSDN endeavors to learn as much as it can about the correspondence between attributes and visual features, but it still construct some wrong relational pairs. Our method preserves the features that correspond correctly, e.g., \emph{black\_wing, black\_forehead, grey\_bill}. Meanwhile, features that are not originally learned are successfully enhanced or captured by our scheme, e.g., \emph{black\_eye, solid\_breast, solid\_belly, black\_underparts}. It illustrates that our method effectively maintains the original valuable knowledge and guides the model to mine more attribute-related features.

\subsection{t-SNE Analysis to Features}

We utilize the t-SNE map on CUB to study the kept and learned features. The experimental results are shown in Fig. \ref{fig:tsne}, where $10$ classes are randomly selected. From the image, we can observe that our method is effective for both seen and unseen classes. Specifically, the features extracted by our method possess more obvious differentiation, e.g., \emph{green, purple, yellowgreen} in Fig. \ref{fig:tsne} (b) and \emph{yellow, cyan, pink} in Fig. \ref{fig:tsne} (d). It demonstrates that the proposed AGL module effectively captures the connection between features and attributes, and thus learns the attribute-related knowledge and transfers it to the unseen domain.

%% file: sec/5_conclusion.tex
\section{Conclusion}
\label{sec:conclusion}

In this paper, we analyze that existing ZSL/GZSL methods are limited by the incompleteness of the extracted features. Such incompleteness stems from the problem of domain bias between the pre-trained model and the downstream tasks. Fine-tuning serves as a simple approach to address this problem, while can introduce catastrophic forgetting and seen class biased overfitting. To address these issues, we present a novel Attribute-Aware Representation Rectification framework, dubbed $\mathbf{(AR)^{2}}$, to refine the learned features while, at the same time, maintaining the original valuable features. Our approach consists of two main modules, i.e., Unseen-Aware Distillation (UAD) and Attribute-Guided Learning (AGL), which dominate the work of keeping old knowledge and learning effective new knowledge, respectively. Through extensive experimental analysis, we show that our method can effectively improve the model's recognition performance in ZSL/GZSL tasks.